\documentclass[conference]{IEEEtran}
\ifCLASSINFOpdf
  \usepackage[pdftex]{graphicx}
  \graphicspath{{.}}
  \DeclareGraphicsExtensions{.pdf,.jpeg,.png}
\else
  \usepackage[dvips]{graphicx}
  \graphicspath{{../eps/}}
  \DeclareGraphicsExtensions{.eps}
\fi
\usepackage[cmex10]{amsmath}
\DeclareMathOperator{\sgn}{sgn}
\usepackage{algorithmicx}
\usepackage{algorithm}
\usepackage{algpseudocode}
\usepackage{caption} 
\begin{document}
%
\title{Adapting Resilient Propagation for Deep Learning}

\author{\IEEEauthorblockN{Alan Mosca}
\IEEEauthorblockA{Department of Computer Science and\\Information Systems\\
Birkbeck, University of London\\
Malet Street, London WC1E 7HX - United Kingdom\\
Email: a.mosca@dcs.bbk.ac.uk}
\and
\IEEEauthorblockN{George D. Magoulas}
\IEEEauthorblockA{Department of Computer Science and\\Information Systems\\
Birkbeck, University of London\\
Malet Street, London WC1E 7HX - United Kingdom\\
Email: gmagoulas@dcs.bbk.ac.uk}}


%


\maketitle

\begin{abstract}
The Resilient Propagation (Rprop) algorithm has been very popular
for backpropagation training of multilayer feed-forward neural networks in
various applications. The standard Rprop however encounters difficulties in
the context of deep neural networks as typically happens with
gradient-based learning algorithms. In this paper, we propose a
modification of the Rprop that combines standard Rprop steps with a special
drop out technique. We apply the method for training Deep Neural Networks
as standalone components and in ensemble formulations. Results on the MNIST
dataset show that the proposed modification alleviates standard Rprop's
problems demonstrating improved learning speed and accuracy.
\end{abstract}


%
\IEEEpeerreviewmaketitle

\section{Introduction}
Deep Learning techniques have generated many of the state-of-the-art
models~\cite{wan2013regularization,ciresan2012multi,ciresan2010deep} that
reached impressive results on benchmark datasets like
MNIST~\cite{mnistlecun}.  Such models are usually trained with variations of the
standard Backpropagation method, with stochastic gradient descent (SGD). In the
field of shallow neural networks, there have been several developments to
training algorithms that have sped up
convergence~\cite{riedmiller93,anastasiadismagoulas05}. This paper aims to
bridge the gap between the field of Deep Learning and these advanced training
methods, by combining Resilient Propagation (Rprop)~\cite{riedmiller93},
Dropout~\cite{hintondropout12} and Deep Neural Networks Ensembles.

\subsection{Rprop}
The Resilient Propagation~\cite{riedmiller93} weight update rule was initially
introduced as a possible solution to the ``vanishing gradients'' problem: as the
depth and complexity of an artificial neural network increase, the gradient
propagated backwards by the standard SGD backpropagation becomes increasingly
smaller, leading to negligible weight updates, which slow down training
considerably.  Rprop solves this problem by using a fixed update value
$\delta_{ij}$, which is increased or decreased multiplicatively at each
iteration by an asymmetric factor $\eta_+$ and $\eta_-$ respectively, depending
on whether the gradient with respect to $w_{ij}$ has changed sign between two
iterations or not. This ``backtracking'' allows Rprop to still converge to a
local minima, but the acceleration provided by the multiplicative factor
$\eta_+$ helps it skip over flat regions much more quickly. To avoid double
punishment when in the backtracking phase, Rprop artificially forces the
gradient product to be $0$, so that the following iteration is skipped. An
illustration of Rprop can be found in Algorithm~\ref{Alg:Rprop}.

\begin{algorithm}
  \caption{Rprop}
  \label{Alg:Rprop}
  \begin{algorithmic}[1]
    \State $\eta_+ = 1.2$, $\eta_- = 0.5$, $\Delta_{max} = 50$, $\Delta_{min} = 10^{-6}$
    \State pick $\Delta_{ij}(0)$
    \State $\Delta w_{ij}(0) = 
        - \sgn\frac{\partial E(0)}{\partial w_{ij}} \cdot \Delta_{ij}(0)$
    \ForAll{$t \in [1..T]$}
      \If{$ \frac{\partial E(t)}{\partial w_{ij}} \cdot
                \frac{\partial E(t-1)}{\partial w_{ij}} > 0 $}
            \State $\Delta_{ij}(t) = \min\{\Delta_{ij}(t-1) \cdot
                \eta_+, \Delta_{max}\}$
        \State $\Delta w_{ij}(t) = - \sgn \frac{\partial E(t)}{\partial w_{ij}}
                \cdot \Delta_{ij}(t)$
            
        \State $w_{ij}(t+1) = w_{ij}(t) + \Delta w_{ij}(t)$
            
        \State $\frac{\partial E(t-1)}{\partial w_{ij}} = \frac{\partial
          E(t)}{\partial w_{ij}}$
      \ElsIf{$ \frac{\partial E(t)}{\partial w_{ij}} \cdot
               \frac{\partial E(t-1)}{\partial w_{ij}} < 0 $}
        \State $\Delta_{ij}(t) = \max\{\Delta_{ij}(t-1) \cdot
                                    \eta_-, \Delta_{min}\}$
        \State $\frac{\partial E(t-1)}{\partial w_{ij}} = 0$
                        
      \Else
      \State $\Delta w_{ij}(t) = - \sgn \frac{\partial E(t)}{\partial w_{ij}}
                                \cdot \Delta_{ij}(t)$
          \State $w_{ij}(t+1) = w_{ij}(t) + \Delta w_{ij}(t)$
        \State $\frac{\partial E(t-1)}{\partial w_{ij}} = \frac{\partial
          E(t)}{\partial w_{ij}}$
      \EndIf
    \EndFor
  \end{algorithmic}
\end{algorithm}

\subsection{Dropout}
Dropout~\cite{hintondropout12} is a regularisation method by which only a
random selection of nodes in the network is updated during each training
iteration, but at the final evaluation stage the whole network is used. The
selection is performed by sampling a \emph{dropout mask} $Dm$ from a Bernoulli
distribution with $P(muted_{i}) = D_r$, where $P(muted_{i})$ is the
probability of node $i$ being muted during the weight update step of
backpropagation, and $D_r$ is the \emph{dropout rate}, which is usually $0.5$
for the middle layers, $0.2$ or $0$ for the input layers, and $0$ for the
output layer. For convenience this dropout mask is represented as a weight
binary matrix $D \in \{0,1\}^{M \times N}$, covering all the weights in the
network that can be used to multiply the weight-space of the network to obtain
what is called a \emph{thinned} network, for the current training iteration,
where each weight $w_{ij}$ is zeroed out based on the probability of its parent
node $i$ being muted.

The remainder of this paper is structured as follows:
\begin{itemize}
    \item In section \ref{Sec:RPROPDROP} we explain why using Dropout causes an
        incompatibility with Rprop, and propose a modification to solve the
        issue.
    \item In section \ref{Sec:MNIST} we show experimental results using the
        MNIST dataset, first to highlight how Rprop is able to converge much
        more quickly during the initial epochs, and then use this to speed up
        the training of a Stacked Ensemble.
    \item Finally in section \ref{Sec:Conclusion}, we look at how this work can
        be extended with further evaluation and development.
\end{itemize}

\section{Rprop and Dropout}
\label{Sec:RPROPDROP}
In this section we explain the \emph{zero gradient problem}, and propose a
solution by adapting the Rprop algorithm to be aware of Dropout.

\subsection{The zero-gradient problem}
In order to avoid double punishment when there is a change of sign in the
gradient, Rprop artificially sets the gradient product associated with weight
$ij$ for the next iteration to $\frac{\partial E_t}{\partial w_{ij}} \cdot
\frac{\partial E_{t+1}}{\partial w_{ij}} = 0$.  This condition is checked during
the following iteration, and if true no updates to the weights $w_{ij}$ or the
learning rate $\Delta_{ij}$ are performed.

Using the zero-valued gradient product as an indication to skip an iteration is
acceptable in normal gradient descent because the only other occurrence of this
would be when learning has terminated.
When Dropout is introduced, an additional number of events can produce these
zero values:
\begin{itemize}
  \item When neuron $i$ is skipped, the dropout mask for all weights $w_{ij}$
        going to the layer above has a value of $0$
  \item When neuron $j$ in the layer above is skipped, the gradient propagated
        back to all the weights $w_{ij}$ is also $0$
\end{itemize}
These additional zero-gradient events force additional skipped training
iterations and missed learning rate adaptations that slow down the training
unnecessarily.

\subsection{Adaptations to Rprop}
By making Rprop aware of the dropout mask $Dm$, we are able to distinguish
whether a zero-gradient event occurs as a signal to skip the next weight update
or whether it occurs for a different reason, and therefore $w$ and $\Delta$
updates should be allowed. The new version of the Rprop update rule for each
weight $ij$ is shown in Algorithm~\ref{Alg:RpropDeep}. We use $t$ to indicate
the current training example, $t-1$ for the previous training example, $t+1$
for the next training example, and where a value with $(0)$ appears, it is
intended to be the initial value. All other notation is the same as used in the
original Rprop:
\begin{itemize}
    \item $E(t)$ is the error function (in this case negative log likelihood)
    \item $\Delta_{ij}(t)$ is the current update value for weight at index $ij$
    \item $\Delta w_{ij}(t)$ is the current weight update value for index $ij$
\end{itemize}

\begin{algorithm}
  \caption{Rprop adapted for Dropout}
  \label{Alg:RpropDeep}
  \begin{algorithmic}[1]
    \State $\eta_+ = 1.2$, $\eta_- = 0.5$, $\Delta_{max} = 50$, $\Delta_{min} = 10^{-6}$
    \State pick $\Delta_{ij}(0)$
    \State $\Delta w_{ij}(0) = 
        - \sgn\frac{\partial E(0)}{\partial w_{ij}} \cdot \Delta_{ij}(0)$
    \ForAll{$t \in [1..T]$}
      \If{$Dm_{ij} = 0$} \label{Lst:FirstAdaptation}
        \State $\Delta_{ij}(t) = \Delta_{ij}(t-1)$
        \State $\Delta w_{ij}(t) = 0$
      \Else
      \If{$ \frac{\partial E(t)}{\partial w_{ij}} \cdot
                \frac{\partial E(t-1)}{\partial w_{ij}} > 0 $}
        \State $\Delta_{ij}(t) = \min\{\Delta_{ij}(t-1) \cdot
                                    \eta_+, \Delta_{max}\}$
                        
        \State $\Delta w_{ij}(t) = - \sgn \frac{\partial E(t)}{\partial w_{ij}}
                \cdot \Delta_{ij}(t)$
            
        \State $w_{ij}(t+1) = w_{ij}(t) + \Delta w_{ij}(t)$
            
        \State $\frac{\partial E(t-1)}{\partial w_{ij}} = \frac{\partial
          E(t)}{\partial w_{ij}}$
      \ElsIf{$ \frac{\partial E(t)}{\partial w_{ij}} \cdot
               \frac{\partial E(t-1)}{\partial w_{ij}} < 0 $}
        \State $\Delta_{ij}(t) = \max\{\Delta_{ij}(t-1) \cdot
                                    \eta_-, \Delta_{min}\}$
        \State $\frac{\partial E(t-1)}{\partial w_{ij}} = 0$
                        
      \Else
        \If{$\frac{\partial E(t-1)}{\partial w_{ij}} = 0$}
        \label{Lst:SecondAdaptation}
        \State $\Delta w_{ij}(t) = - \sgn \frac{\partial E(t)}{\partial w_{ij}}
                                \cdot \Delta_{ij}(t)$
          \State $w_{ij}(t+1) = w_{ij}(t) + \Delta w_{ij}(t)$
        \Else
          \State $\Delta_{ij}(t) = \Delta_{ij}(t-1)$
          \State $\Delta w_{ij}(t) = 0$
        \EndIf
      \EndIf
      \EndIf
    \EndFor
  \end{algorithmic}
\end{algorithm}

In particular, the conditions at line~\ref{Lst:FirstAdaptation} and
line~\ref{Lst:SecondAdaptation} are providing the necessary protection from the
additional zero-gradients, and implementing correctly the recipe prescribed by
Dropout, by completely skipping every weight for which $Dm_{ij} = 0$ (which
means that neuron $j$ was dropped out and therefore the gradient will
necessarily be $0$. We expect
that this methodolgy can be extended to other variants of Rprop, such as, but
not limited to, iRprop+~\cite{igel2000improving} and
JRprop~\cite{anastasiadismagoulas05}.

\section{Evaluating on MNIST}
\label{Sec:MNIST}
In this section we describe an initial evaluation of performance on the MNIST
dataset. For all experiments we use a Deep Neural Network (DNN) with five middle
layers, of $2500,2000,1500,1000,500$ neurons respectively, and a dropout rate
$Dr_{mid} = 0.5$ for the middle layers and no Dropout on the inputs. The dropout
rate has been shown to be an optimal choice for the MNIST dataset in
\cite{srivastava2014dropout}. A similar architecture has been used to produce
state-of-the-art results~\cite{ciresan2010deep}, however the authors used the
entire training set for validation, and graphical transformations of said set
for training.  These added transformations have led to a ``virtually infinite''
training set size, whereby at every epoch, a new training set is generated, and
a much larger validation set of the original $60000$ images. The test set
remains the original $10000$ image test set. An explanation of these
transformations is provided in ~\cite{simard2003best}, which also confirms that:
\begin{quotation}
    ``The most important practice is getting a training set as large as
    possible: we expand the training set by adding a new form of distorted
    data''
\end{quotation}
We therefore attribute these big improvements to the transformations applied, and
have not found it a primary goal to replicate these additional transformations
to obtain the state-of-the-art results and instead focused on utilising the
untransformed dataset, using $50000$ images for training, $10000$ for validation
and $10000$ for testing.
Subsequently, we performed a search using the validation set as an indicator to
find the optimal hyperparameters of the modified version of Rprop. We found that
the best results were reached with $\eta_+ = 0.01$, $\eta_- = 0.1$,
$\Delta_{max} = 5$ and $\Delta_{min} = 10^{-3}$.
We trained all models to the maximum of $2000$ allowed epochs, and measured the
error on the validation set at every epoch, so that it could be used to select
the model to be applied to the test set. We also measured the time it took to
reach the best validation error, and report its approximate magnitude, to use as
a comparison of orders of magnitude. The results presented are an average of $5$
repeated runs, limited to a maximum of $2000$ training epochs.

\subsection{Compared to SGD}
\begin{figure}[h]
\centering
\includegraphics[width=\linewidth]{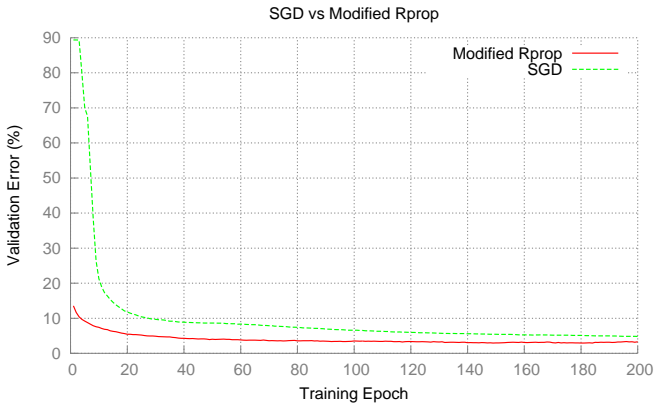}
\caption{Validation Error - SGD vs Mod. Rprop}
\label{Fig:RpropVsSGD}
\end{figure}

From the results in Table~\ref{Tab:Results} we see that the modified version of
Rprop is able to start-up much quicker and reaches an error value that is close
to the minimum much more quickly. SGD reaches a higher error value, and after a
much longer time. Although the overall error improvement is significant, the
speed gain from using Rprop is more appealing because it allows to save a large
number of iterations that could be used for improving the model in different
ways. Rprop obtains its best validation error after only  $35$ epochs, whilst
SGD reached the minimum after $473$. An illustration of the first $200$ epochs
can be seen in Figure~\ref{Fig:RpropVsSGD}.

\begin{table}[h]
  \centering
  \begin{tabular}{ l | l l l l l }
    \hline
    Method    & Min Val Err & Epochs & Time    & Test Err & $1^{st}$ Epoch \\
    \hline
    SGD       & 2.85\%      & 1763   & 320 min & 3.50\%   & 88.65\% \\
    Rprop     & 3.03\%      & 105    & 25 min  & 3.53\%   & 12.81\% \\
    Mod Rprop & 2.57\%      & 35     & 10 min  & 3.49\%   & 13.54\% \\
    \hline
  \end{tabular}
  \caption{Simulation results}
  \label{Tab:Results}
\end{table}

\subsection{Compared to unmodified Rprop}
\begin{figure}[h]
\centering
\includegraphics[width=\linewidth]{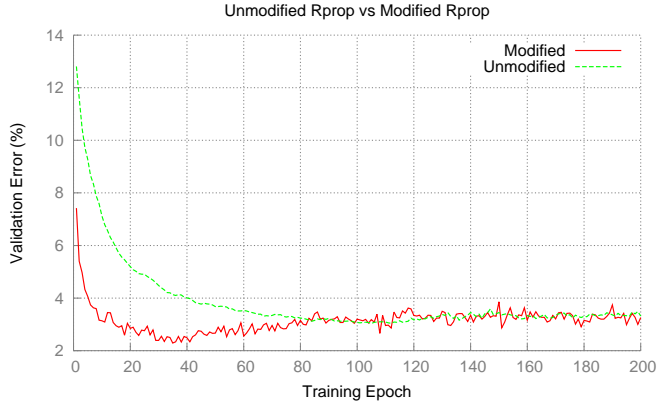}
\caption{Validation Error - Unmod. vs Mod. Rprop}
\label{Fig:NewvsOld}
\end{figure}

We can see from Figure~\ref{Fig:NewvsOld} that the modified version
of Rprop has a faster start-up than the unmodified version, and stays below it
consistently until it reaches its minimum.
Also, the unmodified version does not reach the same final error as the modified
version, and starts overtraining much sooner, and does not reach a better error
than SGD. Table~\ref{Tab:Results} shows with more detail how the performance of
the two methods compares over the first $200$ epochs.

\subsection{Using Modified Rprop to speed up training of Deep Learning Ensembles}
The increase in speed of convergence can make it practical to produce Ensembles
of Deep Neural Networks, as the time to train each member DNN is considerably
reduced without undertraining the network. We have been able to train these
Ensembles in less than 12 hours in total on a single-GPU, single-CPU desktop
system \footnote{We used a Nvidia GTX-770 graphics card on a core i5 processor,
programmed with Theano in python}.  We have trained different Ensemble types,
and we report the final results in Table~\ref{Tab:Ensembles}.
The methods used are Bagging~\cite{breiman96} and Stacking~\cite{wolpert92}, with
$3$ and $10$ member DNNs. Each member was trained for a maximum of $50$ epochs.
\begin{itemize}
    \item Bagging is an ensemble method by which several different training sets
        are created by random resampling of the original training set, and each
        of these are used to train a new classifier. The entire set of trained
        classifiers is usually then aggregated by taking an average or a
        majority vote to reach a single classification decision.
    \item Stacking is an ensemble method by which the different classifiers are
        aggregated using an additional learning algorithm that uses the inputs
        of these \emph{first-space} classifiers to learn information about how to
        reach a better classification result. This additional learning algorithm
        is called a \emph{second-space} classifier.
\end{itemize}

In the case of Stacking the final second-space classifier was another DNN with
two middle layers, respectively of size $(200N,100N)$, where $N$ is the number
of DNNs in the Ensemble, trained for a maximum of $200$ epochs with the modified
Rprop.  We used the same original train, validation and test sets for this, and
collected the average over $5$ repeated runs. The results are still not
comparable to what is presented in ~\cite{ciresan2010deep}, which is consistent
with the observations about the importance of the dataset transformations,
however we note that we are able to improve the error in less time it took to
train a single network with SGD. A Wilcoxon signed ranks test shows that the increase in performance
obtained from using the ensembles of size $10$ compared to the ensemble of size
$3$ is significant, at the $98\%$ confidence level.

\begin{table}[h]
  \centering
  \begin{tabular}{ l l  | l l l l }
    \hline
    Method    & Size & Test Err & Time    \\
    \hline
    Bagging   & $3$  & $2.56\%$ & 35 min  \\
    Bagging   & $10$ & $2.13\%$ & 128 min \\
    Stacking  & $3$  & $2.48\%$ & 39 min \\
    Stacking  & $10$ & $2.19\%$ & 145 min \\
    \hline
  \end{tabular}
  \caption{Ensemble performance}
  \label{Tab:Ensembles}
\end{table}

\section{Conclusions and Future Work}
\label{Sec:Conclusion}
We have highlighted that many training methods that have been used in shallow
learning may be adapted for use in Deep Learning. We have looked at Rprop and
how the appearance of \emph{zero-gradients} during the training as a side effect
of Dropout poses a challenge to learning, and proposed a solution which allows
Rprop to train DNNs to a better error, and still be much faster than standard
SGD backpropagation.

We then showed that this increase in training speed can be used to train
effectively an Ensemble of DNNs on a commodity desktop system, and reap the
added benefits of Ensemble methods in less time than it would take to train a
Deep Neural Network with SGD.

It remains to be assessed in further work whether this improved methodology
would lead to a new state-of-the-art error when applying the pre-training and
dataset enhancements that have been used in other methods, and how the
improvements to Rprop can be ported to its numerous variants.

\section*{Acknowledgement}
The authors would like to thank the School of Business, Economics and
Informatics, Birkbeck College, University of London, for the grant received to
support this research.


\bibliographystyle{IEEEtran}
\bibliography{../../biblio}
\end{document}